\documentclass[]{bytedance_seed}

\usepackage[toc,page,header]{appendix}

\usepackage{minitoc}
\usepackage{amsfonts}
\usepackage{amssymb}
\usepackage{algorithm}
\usepackage{algpseudocode}

\title{ SplitMeanFlow: Interval Splitting Consistency in Few-Step Generative Modeling  }

\author[*, ]{Yi Guo}
\author[*]{Wei Wang}
\author[*]{Zhihang Yuan}
\author{Rong Cao}
\author{Kuan Chen}
\author{Zhengyang Chen}
\author{Yuanyuan Huo}
\author{Yang Zhang}
\author{Yuping Wang}
\author{Shouda Liu}
\author{Yuxuan Wang}
\affiliation{ByteDance Seed}
\contribution[*]{Equal Contribution}
\abstract{
Generative models like Flow Matching have achieved state-of-the-art performance but are often hindered by a computationally expensive iterative sampling process. To address this, recent work has focused on few-step or one-step generation by learning the \textit{average velocity} field, which directly maps noise to data. MeanFlow, a leading method in this area, learns this field by enforcing a differential identity that connects the average and instantaneous velocities. In this work, we argue that this differential formulation is a limiting special case of a more fundamental principle. We return to the first principles of average velocity and leverage the additivity property of definite integrals. This leads us to derive a novel, purely algebraic identity we term \textbf{Interval Splitting Consistency}. This identity establishes a self-referential relationship for the average velocity field across different time intervals without resorting to any differential operators.
Based on this principle, we introduce \textbf{SplitMeanFlow}, a new training framework that enforces this algebraic consistency directly as a learning objective. We formally prove that the differential identity at the core of MeanFlow is recovered by taking the limit of our algebraic consistency as the interval split becomes infinitesimal. This establishes SplitMeanFlow as a direct and more general foundation for learning average velocity fields. From a practical standpoint, our algebraic approach is significantly more efficient, as it eliminates the need for JVP computations, resulting in simpler implementation, more stable training, and broader hardware compatibility. 
One-step and two-step SplitMeanFlow models have been successfully deployed in large-scale speech synthesis products (such as Doubao), achieving speedups of 20$\times$.
}
\date{\today}
\begin{document}
\maketitle

%不需要目录就注释掉 注意目录不要和第一页放在一块 要有\newpage
%\newpage
%\tableofcontents
%\newpage

\section{Introduction}

The field of generative modeling has witnessed remarkable progress, with methods like Diffusion Models~\cite{diffusion} and Flow Matching~\cite{liu2022flow,scoresde} setting new standards in generating high-fidelity samples across various domains, including images~\cite{rombach2022high,esser2024scaling}, video~\cite{kong2024hunyuanvideo,bar2024lumiere} and audio~\cite{guan2024lafma,tian2025audiox}. Despite their power, the practical utility of these models is often hindered by a significant computational bottleneck: their reliance on an iterative sampling process that typically requires tens or even hundreds of neural network inferences. This substantial computational cost poses a major challenge for real-world applications, particularly in resource-constrained or latency-sensitive environments. Consequently, a vibrant research area has emerged, focusing on developing ``few-step'' or even ``one-step'' generative models to drastically reduce this sampling overhead.

In response to this challenge, several innovative approaches have been proposed. Consistency Models~\cite{cm,luo2023latent}, for instance, introduced a novel training paradigm by enforcing output consistency for points along the same trajectory, achieving promising results in few-step generation. Building on this momentum, MeanFlow~\cite{meanflow} offered a profound and physically intuitive insight: for large-step generation, it is more effective to directly model the \textit{average velocity} along the entire path connecting noise to data, rather than the \textit{instantaneous velocity} at each point. This conceptual shift from a local to a global perspective is inherently better suited for few-step, large-stride predictions and has led to state-of-the-art performance.

The success of MeanFlow~\cite{meanflow} prompts a deeper investigation into its underlying mechanism. How does it effectively learn the average velocity field? A closer look reveals that its implementation does not directly compute the integral from its definition. Instead, it ingeniously leverages a differential identity that connects the average velocity \( u \) and the instantaneous velocity \( v \): \( u = v - (t-r)du/dt \). While remarkably effective, this differential formulation raises a fundamental question: Does this approach, which relies on derivatives, fully capture the intrinsic nature of average velocity? Or is it a clever but potentially limited perspective?

We argue for the latter. In this work, we advocate for a return to the first principles of average velocity: its definition as the integral of instantaneous velocity over a time interval, \( u \propto \int v d\tau \). 
% Instead of approximating this integral relationship with derivatives, we embrace its fundamental algebraic structure.
The cornerstone of our approach is the additivity property of definite integrals: for any intermediate time \( s \in [r, t] \), the integral over \( [r, t] \) is the sum of the integrals over \( [r, s] \) and \( [s, t] \) (i.e., \( \int_{r}^{t} = \int_{r}^{s} + \int_{s}^{t} \)).
By substituting the definition of displacement, \( (t-r)u(z_t, r, t) = \int_{r}^{t} v d\tau \), this property translates into an exact algebraic equivalence. We derive a novel, purely algebraic identity that we term \textbf{Interval Splitting Consistency}:
\begin{equation}
    (t-r)u(z_t,r,t) = (s-r)u(z_s,r,s) + (t-s)u(z_t,s,t).
\end{equation}
This identity governs the intrinsic relationship between average velocities across different time intervals without resorting to any differential operators. Based on this principle, we introduce \textbf{SplitMeanFlow}, a new framework for learning the average velocity field by directly enforcing this consistency as a training objective.

Crucially, this algebraic formulation is not merely an alternative but a more general foundation. We formally demonstrate that the differential identity at the core of MeanFlow is recovered by taking the limit of our Interval Splitting Consistency as the splitting point \( s \) approaches the endpoint \( t \). In this limit, our algebraic relation gracefully collapses into the differential form, revealing that MeanFlow's training objective is, in fact, a special case of our more fundamental consistency principle. This connection establishes SplitMeanFlow as a direct generalization of MeanFlow, offering a more comprehensive and robust framework for learning average velocity fields.

Our contributions are summarized as follows:
\begin{itemize}
    \item \textbf{A General and Principled Framework.} We introduce SplitMeanFlow, a framework grounded in an algebraic \textit{Interval Splitting Consistency}. We prove that the differential identity of MeanFlow~\cite{meanflow} is a limiting special case of our formulation, establishing our method's theoretical generality.
    \item \textbf{JVP-Free, Efficient, and Simple.} Our approach eliminates the need for expensive Jacobian-vector product (JVP) computations, leading to more stable training,
    % significantly faster training, 
    broader hardware compatibility, and a simpler implementation.
    % \item \textbf{State-of-the-Art Performance.} SplitMeanFlow achieves state-of-the-art results in few-step generation, outperforming previous methods while being more efficient.
    \item \textbf{Proven Industrial Impact.} Our method has been successfully deployed in large-scale industrial products, DouBao, demonstrating its practical value and robustness.
\end{itemize}

\section{Related Work}

\subsection{Diffusion Models and Flow Matching}
Diffusion models \cite{diffusion, song2019score,ho2020denoising,scoresde,nichol2021improved,rombach2021high,peebles2023scalable} have achieved impressive results across various generative tasks by transforming noise into data through iterative denoising. Despite their success, these models often require hundreds of sampling steps, making inference computationally expensive and limiting their applicability in real-time settings~\cite{yang2023diffusion}.
To address this inefficiency, flow matching \cite{lipman2022flow,Karras2022edm,albergo2023stochastic} has been proposed as an alternative framework that directly learns a time-dependent velocity field to match the probability flow of a diffusion process. Flow Matching offers advantages such as faster inference and empirical performance gains.
Building on this framework, recent variants such as Rectified Flow \cite{liu2022flow} aim to improve training stability and convergence by adjusting the reference interpolation path. However, despite these improvements, most FM-based approaches still rely on multi-step integration during sampling. To reduce the computational burden of iterative sampling, recent research has focused on few-step generative models that aim to accelerate inference while preserving the high sample quality of diffusion-based methods.

\subsection{Few-Step Generative Models}
\textbf{Consistency Models.}
Consistency models~\cite{cm,heek2024multistep,lu2024simplifying} have been developed to achieve few-step generation for visual~\cite{luo2023latent,oertell2024rl} and audio~\cite{fei2024music,liu2024audiolcm}. These methods enforce self-consistency by requiring that predictions remain invariant under repeated model application or temporal interpolation across varying noise levels. Such constraints encourage the generative trajectory to become coherent and predictable, thereby allowing accurate approximation with substantially fewer steps. Despite their empirical effectiveness, these consistency constraints are generally heuristic in nature, introduced as external regularization without explicit theoretical grounding. 

% \textbf{Shortcut Models.}
% Shortcut models~\cite{frans2024one} present an alternative strategy by directly learning mappings from noise to data, leveraging intermediate states along the diffusion trajectory as supervision signals. Typically trained through teacher-student distillation or similar heuristic approaches, shortcut models effectively shorten the generative path, thus significantly reducing the number of inference steps required. Nevertheless, this class of methods also relies heavily on empirical design, employing consistency objectives defined only at specific intermediate points, such as trajectory midpoints. 
% This selective enforcement of consistency may limit their theoretical clarity and general applicability, motivating further research toward more principled frameworks.

\textbf{MeanFlow} MeanFlow~\cite{meanflow} presents a principled framework for one-step generative modeling by introducing the concept of \textit{average velocity}, defined as the displacement over a time interval divided by its duration. 
In contrast to Flow Matching, which models instantaneous velocity at each time step, MeanFlow adopts average velocity as the learning target. It further derives an analytic relation, termed the \textit{MeanFlow Identity}, that connects the average and instantaneous velocities via a time derivative.
This formulation offers a well-grounded training objective that avoids heuristic consistency constraints and provides a clear physical interpretation. 
% The method is self-contained and does not rely on score estimation, pretraining, or distillation.
% Nevertheless, a key limitation of MeanFlow lies in its reliance on computing \textbf{Jacobian-vector products (JVPs)} to evaluate the time derivative. While supported by modern autodiff frameworks, JVPs introduce additional computational overhead and may be incompatible with custom CUDA operators that do not implement JVP interfaces, posing challenges for scalability and integration.

%\subsection{Hello World}
\section{Preliminary: Flow Matching}

To properly contextualize our proposed SplitMeanFlow, we first revisit the foundational principles of Flow Matching, the framework upon which our work is built. Flow Matching offers a powerful and intuitive paradigm for generative modeling, designed to learn a velocity field that transports samples from a simple prior distribution (e.g., a Gaussian) to a complex target data distribution. Both the preceding MeanFlow model and our SplitMeanFlow are fundamentally grounded in the core mechanics of this approach.

\subsection{Flow Paths and Instantaneous Velocity}

The central idea of Flow Matching~\cite{diffusion, song2019score,ho2020denoising,scoresde,nichol2021improved,rombach2021high,peebles2023scalable} is to define a continuous-time \textit{flow path}, denoted by \(z_t\), that connects a prior sample \(\epsilon \sim p_{\text{prior}}(\epsilon)\) to a data sample \(x \sim p_{\text{data}}(x)\). This path is typically parameterized over the time interval \(t \in [0, 1]\) as:
\begin{equation}
    z_t = a_t x + b_t \epsilon,
\end{equation}
where \(a_t\) and \(b_t\) are predefined scalar schedules satisfying boundary conditions such as \(a_0=1, b_0=0\) and \(a_1=0, b_1=1\). A common and simple choice is the linear schedule \(a_t = 1-t\) and \(b_t = t\), which defines a straight-line trajectory from \(x\) at \(t=0\) to \(\epsilon\) at \(t=1\).

Associated with this path is an \textit{instantaneous velocity field} \(v_t\), defined as the time derivative of the path:
\begin{equation}
    v_t = \frac{d z_t}{dt} = a'_t x + b'_t \epsilon.
\end{equation}
Since this velocity is defined conditioned on a specific data sample \(x\), it is referred to as the \textbf{conditional velocity}, denoted \(v_t(z_t | x)\). For the linear schedule mentioned above, the conditional velocity takes the simple form \(v_t = \epsilon - x\).

\subsection{Conditional Flow Matching Loss}

In practice, any given point \(z_t\) on a trajectory could have been generated by numerous different \((x, \epsilon)\) pairs. The ultimate goal of a generative model is therefore not to learn any single conditional velocity, but rather the expectation over all possibilities, known as the \textbf{marginal velocity} \(v(z_t, t) = \mathbb{E}[v_t(z_t|x) | z_t]\). However, directly computing and optimizing a loss against this marginal velocity is intractable.

To circumvent this challenge, Flow Matching~\cite{lipman2022flow,Karras2022edm,albergo2023stochastic} introduces an elegant and practical objective: the \textbf{Conditional Flow Matching (CFM) loss}. This objective trains a neural network \(v_\theta\) by minimizing the discrepancy between its prediction and the easily computable conditional velocity. The loss function is formulated as:
\begin{equation}
    \mathcal{L}_{\text{CFM}}(\theta) = \mathbb{E}_{t, x, \epsilon} \left\| v_\theta(z_t, t) - (a'_t x + b'_t \epsilon) \right\|^2.
\end{equation}
It has been shown that minimizing this conditional loss is equivalent to minimizing the loss with respect to the true marginal velocity field. By optimizing this objective, the network \(v_\theta\) effectively learns the vector field that governs the transformation of the entire distribution.

Once the model \(v_\theta\) is trained, new samples can be generated by solving the ordinary differential equation (ODE) \( \frac{dz_t}{dt} = v_\theta(z_t, t) \), starting from a prior sample \(z_1 = \epsilon\) and integrating backward in time to \(t=0\). This integration typically requires a numerical ODE solver, which often involves multiple evaluation steps and has motivated the research into more efficient few-step and one-step generation methods.
\section{Method}

\begin{figure}[t] % h:here 当前位置 % b bottom % t top % p 浮动
    \centering
    \includegraphics[width=0.98\textwidth]{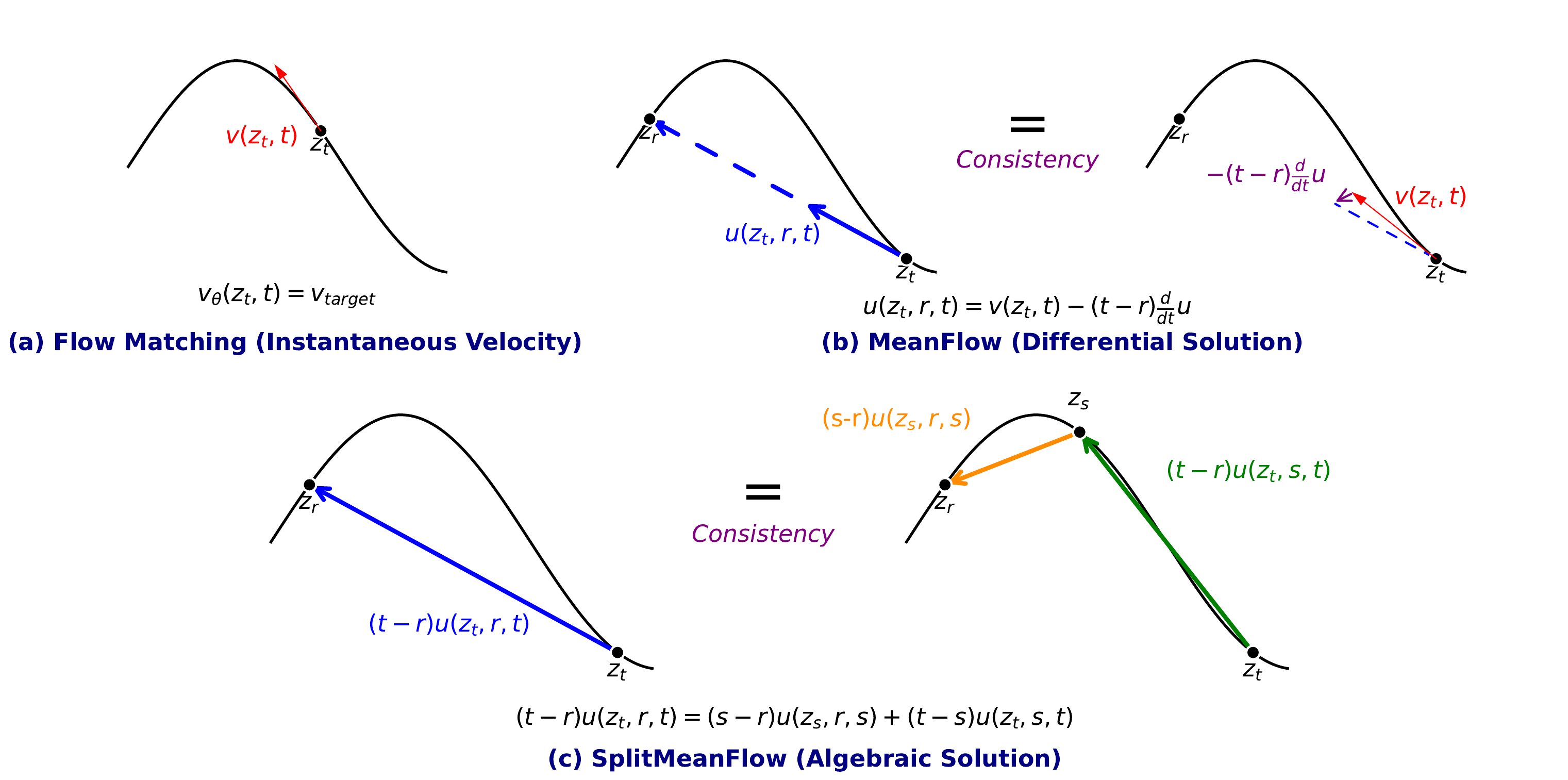} %ims/xx.png
    \caption{Conceptual Comparison of Generative Flow Methods}
    \label{im:conceptual_comparison}
\end{figure}

\subsection{Why Average Velocity Field?}

In the pursuit of generative modeling, a paramount goal is to achieve efficient transformations from a simple prior distribution to a complex data distribution with minimal computational overhead. This objective becomes particularly critical in few-step, and especially one-step, generation scenarios. However, prevailing generative paradigms, such as Diffusion Models and Flow Matching, reveal their inherent limitations when confronted with this extreme efficiency challenge. To understand and surmount this bottleneck, we must return to the foundational question of the learning process: What, precisely, should a model learn to enable both efficient and accurate generation?

\subsubsection{The Imperative: From Instantaneous to Average Velocity}

Conventional generative frameworks are built upon learning an \textbf{instantaneous velocity field}, denoted as \( v(z_t, t) \) (As shown in Figure~\ref{im:conceptual_comparison}a). This field describes the direction and magnitude of change for a data point \( z_t \) at a specific moment in time \( t \). During inference, the model simulates the entire trajectory from a prior sample \( z_1 \) to the target data point \( z_0 \) by numerically integrating this velocity field, effectively solving the ordinary differential equation (ODE):

\begin{equation}
    z_0 = z_1 - \int_{0}^{1} v(z_\tau, \tau) d\tau.
\end{equation}

When a sufficient number of function evaluations (NFE) are employed, numerical solvers like the Euler method or higher-order alternatives can accurately approximate this integral, yielding high-quality samples. This is the cornerstone of multi-step generation.

However, this reliance on fine-grained discretization becomes untenable in the quest for ultimate efficiency. As the number of sampling steps is drastically reduced, the error inherent in numerical integration is severely amplified. In the limit of \textbf{one-step generation}, the process collapses to a single Euler step:

\begin{equation}
    z_0 \approx z_1 - 1 \cdot v(z_1, 1).
\end{equation}

This approximation is fundamentally flawed. It equates the average rate of change over the entire interval \( [0, 1] \) with the instantaneous velocity at the terminal point \( t=1 \). This is conceptually akin to calculating a car's average speed over a full journey using only its speed as it crossed the finish line. Given that the instantaneous velocity field \( v \) is typically highly non-linear with respect to time, this crude approximation introduces a significant \textbf{discretization error}, which stands as the primary obstacle to high-quality one-step generation.

Therefore, to achieve superior few-step and one-step generation, we must move beyond the paradigm of integrating instantaneous velocities at inference time. The solution is both direct and compelling: the model must learn to predict a quantity that encapsulates the entire transformation in a single step. This ideal target is the \textbf{average velocity field}, \( u \). By definition, it is formulated to satisfy the exact relation \( z_0 = z_1 - u(z_1, 0, 1) \). By learning \( u \), we shift the burden of integration from the inference phase to the training objective, conceptually eliminating the discretization error by design. Learning the average velocity field is the key to unlocking efficient and accurate generation.

\subsubsection{First Principles: The Integral Definition of Average Velocity}

Having established the necessity of learning the average velocity field \( u \), we must ground our approach in its fundamental mathematical definition—its \textbf{first principle}. The average velocity is, by definition, the integral of the instantaneous velocity over a given time interval. This serves as the axiom for all subsequent derivations.

Formally, the average velocity \( u \) is defined as:

\begin{equation}
    u(z_t, r, t) \triangleq \frac{1}{t - r} \int_{r}^{t} v(z_\tau, \tau) d\tau.
\end{equation}

This equation lucidly defines \( u \) as a ground-truth field, whose properties are intrinsically determined by the underlying instantaneous field \( v \). It represents an ideal, model-agnostic target.

However, this very definition exposes the core technical challenge: the integral is \textbf{intractable} to compute during training. Its evaluation would require knowing the instantaneous velocity \( v(z_\tau, \tau) \) at all intermediate points \( z_\tau \) along an unknown trajectory, rendering it unusable as a direct learning objective.

This brings us to a critical juncture. Given that the average velocity field \( u \) is the ideal learning target, yet its integral definition is computationally intractable, how can we design a tractable and effective objective function to train a neural network \( u_\theta \) to faithfully approximate \( u \)? The answer to this question motivates two distinct methodological paths, which we will explore in the subsequent section.

\subsection{Two Solution Paths: Differential Solution vs. Algebraic Solution}

Confronted with the core challenge of training a neural network to learn the integral-defined average velocity, two technical paths emerge. The first is a ``differential solution,'' exemplified by MeanFlow, which cleverly circumvents the integral. The second, which we propose in this work, is an ``algebraic solution'' that returns to the fundamental nature of the integral itself.

\subsubsection{Path One: The Differential Solution of MeanFlow}

\textbf{Core Idea:} The fundamental approach of MeanFlow~\cite{meanflow} is to transform the intractable problem of integration into a more manageable problem of differentiation. It recognizes the difficulty of directly computing \( u \propto \int v d\tau \) and instead circumvents it by establishing a differential relationship between the average velocity \( u \) and the instantaneous velocity \( v \), leveraging the fundamental theorem of calculus.

\textbf{Derivation and Training:} The method begins with the definition of displacement, \( D(r, t) = (t-r)u(z_t, r, t) = \int_{r}^{t} v(z_\tau, \tau) d\tau \). By taking the total derivative with respect to time \( t \) on both sides of the equation and applying the fundamental theorem of calculus (\( \frac{d}{dt}\int_{r}^{t} f(\tau)d\tau = f(t) \)) along with the product rule, one can derive the following \textbf{MeanFlow Identity}:
\begin{equation}
    u(z_t, r, t) = v(z_t, t) - (t - r) \frac{d}{dt} u(z_t, r, t)
\label{eq:meanflow_identity}
\end{equation}
This identity provides an effective regression target. During training, MeanFlow treats the right-hand side of the equation as the target value \( u_{\text{tgt}} \) and minimizes the discrepancy between the model's prediction \( u_\theta \) and this target. Specifically, the loss function is \( \mathcal{L}(\theta) = \mathbb{E} \left\| u_\theta(z_t, r, t) - \text{sg}(u_{\text{tgt}}) \right\|_2^2 \), where \( \text{sg} \) denotes the stop-gradient operator. The instantaneous velocity \( v(z_t, t) \) in the target can be substituted with the known conditional velocity \( v_t = \epsilon - x \), while the core differential term \( \frac{d}{dt} u \) is computed via the automatic differentiation capabilities of the neural network framework.

\textbf{Essence and Cost:} The essence of MeanFlow is an \textbf{indirect matching in a differential form}. It does not directly compel \( u_\theta \) to fit the result of the integral. Instead, it constrains \( u_\theta \) to satisfy a differential equation that is itself derived from the integral's definition. The efficacy of this method hinges on a critical computational step: the calculation of the total derivative \( \frac{d}{dt} u \). According to the chain rule, \( \frac{d}{dt} u = v \cdot \nabla_z u + \partial_t u \), which necessitates a Jacobian-Vector Product (JVP) operation. 
%The computational complexity of a JVP is comparable to one backward pass (backpropagation). This means that each training step in MeanFlow involves a JVP operation (akin to a forward + backward pass) to construct the target, followed by a second, true backward pass to update the model parameters.

\subsubsection{Path Two: The Algebraic Solution of SplitMeanFlow}

\textbf{Core Idea:} We advocate for a return to the first principles of integration, directly leveraging its most fundamental algebraic property—additivity—rather than taking a detour through differentiation. We posit that the intrinsic structure of the integral itself contains sufficient self-consistency information, obviating the need for external differential operators.

\textbf{Derivation and Training:} 
Our derivation forgoes differential operators and returns to the most fundamental property of the integral itself: its algebraic additivity. This approach allows us to formulate a self-contained and computationally efficient training objective.

The entire derivation is built upon a simple and profound mathematical axiom: the additivity of definite integrals. For any function integrated over a continuous domain, the integral over a whole interval is precisely the sum of the integrals over its sub-intervals. When applied to the instantaneous velocity field \(v\), this axiom states that for any ordered time points \(r \le s \le t\), the following identity holds:
\begin{equation}
\int_{r}^{t} v(z_\tau, \tau) d\tau = \int_{r}^{s} v(z_\tau, \tau) d\tau + \int_{s}^{t} v(z_\tau, \tau) d\tau 
\end{equation}
This equation reveals a fundamental truth about the geometry of the flow: the total displacement of a particle from time \(r\) to \(t\) is equal to the displacement from \(r\) to an intermediate time \(s\), plus the subsequent displacement from \(s\) to \(t\). 

We can now connect this principle to the average velocity \(u\). By recalling the definition of displacement as the product of average velocity and the time duration, \( D(a, b) = (b-a)u(z_b, a, b) = \int_{a}^{b} v(z_\tau, \tau) d\tau \), we can substitute this definition back into the integral additivity equation. This substitution directly translates the additivity of displacements into a purely algebraic relationship concerning the average velocity field itself. This yields what we term the \textbf{Interval Splitting Consistency} identity:
\begin{equation}
    (t-r)u(z_t,r,t) = (s-r)u(z_s,r,s) + (t-s)u(z_t,s,t)
\label{eq:interval_splitting_conssitency}
\end{equation}
This identity is the cornerstone of SplitMeanFlow. It provides a powerful, self-referential constraint on the structure of the average velocity field.

To facilitate training, dividing both sides of Eq.~\ref{eq:interval_splitting_conssitency} by $t-r$, and letting $\lambda=\frac{t-s}{t-r} \in [0, 1]$ (it gives $s=(1-\lambda)t+\lambda r$), the identity is then transformed into:
\begin{equation}
    u(z_t,r,t) = (1-\lambda)u(z_s,r,s) + \lambda u(z_t,s,t)
\label{eq:interval_splitting_conssitency3}
\end{equation}
The identity conveys an intuitive physical meaning: the average velocity over $[r, t]$ is equivalent to a weighted sum of the average velocities over $[r, s]$ and $[s, t]$, where the weighting coefficients are proportional to the lengths of the respective intervals. 
As a difference equation formulation, identity~\ref{eq:interval_splitting_conssitency3} necessitates boundary constraints to prevent degenerate solutions. The essential boundary condition is given by: when $r=t$, $u(z_t, r, t)=v(z_t, t)$. The training process is referred to Algorithm~\ref{alg:splitmeanflow_training} (For brevity, the boundary conditions are omitted here and will be detailed in the experimental section).

% (As shown in Figure~\ref{im:conceptual_comparison}c). 
%Crucially, it relates the model's predictions at different points in time and over different intervals—\(u(z_t,r,t)\), \(u(z_s,r,s)\), and \(u(z_t,s,t)\)—without any reference to the intractable instantaneous velocity \(v\). This forms a perfect self-supervised learning objective. We can train a neural network \(u_\theta\) by simply enforcing that its outputs satisfy this algebraic conservation law. During training, we sample three time points \((r, s, t)\), compute the three corresponding average velocity predictions with \(u_\theta\), and minimize the difference between the left and right sides of the consistency equation (Algorithm~\ref{alg:splitmeanflow_training}).

\begin{algorithm}[H]
    \caption{SplitMeanFlow Training}
    \label{alg:splitmeanflow_training}
    \begin{algorithmic}[1]
        \Require Neural network \(u_\theta\), a batch of data \(x\), optimizer.
        \State Sample time points \(r, t\) such that \(0 \le r \le t \le 1\), and sample $\lambda \sim \mathcal{U}(0,1)$, set $s=(1-\lambda)t+\lambda r$.
        \State Sample prior \(\epsilon \sim \mathcal{N}(0, I)\).
        \State Construct flow path point at time t: \(z_t = (1-t)x + t\epsilon\).
        \State \(u_2 = u_\theta(z_t, s, t)\) 
        % \Comment{Predict velocity for the second interval}
        \State \(z_s = z_t - (t-s) u_2\) 
        % \Comment{Predict intermediate point \(\hat{z}_s\) using the model}
        \State \(u_1 = u_\theta(z_s, r, s)\) 
        \State $target = (1-\lambda)u_1+\lambda u_2$
        % \State \(u = u_\theta(z_t, r, t)\)
        \State $\mathcal{L}=\|u_\theta(z_t, r, t)-\text{sg}(target)\|$ \Comment{\text{sg} is the stop gradient function.}
        \State Update \(\theta\) using the gradient of $\mathcal{L}$.
        % \Comment{Predict velocity for the first interval at the predicted point}
        % \State \(\text{disp}_{\text{total}} = (t-r) \cdot u_{\text{total}}\)
        % \State \(\text{disp}_{\text{split}} = (s-r) \cdot u_{\text{split1}} + (t-s) \cdot u_{\text{split2}}\)
        % \State \(L = \left\| \text{disp}_{\text{total}}- \text{sg}(\text{disp}_{\text{split}}) \right\|_2^2\)
        % \Comment{\text{sg} is the stop gradient function.}
        % \State Update \(\theta\) using the gradient of \(L\).
    \end{algorithmic}
\end{algorithm}

\textbf{Essence and Cost:} The essence of SplitMeanFlow is an \textbf{algebraic self-consistency constraint}. It establishes a fully self-contained training paradigm where the model learns the intrinsic structure of the average velocity field by `supervising itself.` 

\subsection{Theoretical Analysis: The Superiority of the Algebraic Solution}
\label{sec:theoretical_analysis}

Having introduced the algebraic solution of SplitMeanFlow, a critical question arises: why is this path, which returns to first principles, superior to the established differential solution of MeanFlow? 

\subsubsection{Theoretical Generality}

Our central thesis is that the differential identity underpinning MeanFlow is not an independent law parallel to our algebraic consistency, but rather a limiting special case derived from it. This mathematical relationship establishes SplitMeanFlow as a more fundamental and general framework.

Our proof begins with the cornerstone of SplitMeanFlow, the Interval Splitting Consistency identity (Eq~\ref{eq:interval_splitting_conssitency}).
To reveal its connection to MeanFlow, we first rearrange this equation algebraically:
\begin{equation}
    \frac{(t-r)u(z_t,r,t) - (s-r)u(z_s,r,s)}{t-s} = u(z_t,s,t).
    \label{eq:rearranged_consistency}
\end{equation}
This form already resembles the definition of a derivative. We now investigate its behavior in the limit as the splitting point \(s\) approaches the right endpoint \(t\), i.e., \(s \to t\).

\begin{enumerate}
    \item \textbf{Analyzing the Right-Hand Side (RHS):} As \(s \to t\), the length of the time interval \([s, t]\) for the average velocity \(u(z_t, s, t)\) approaches zero. By its definition, \(u(z_t, s, t) = \frac{1}{t-s}\int_{s}^{t} v(z_\tau, \tau) d\tau\), which converges to the instantaneous velocity at that point. Thus:
    \begin{equation}
        \lim_{s \to t} u(z_t, s, t) = v(z_t, t).
    \end{equation}

    \item \textbf{Analyzing the Left-Hand Side (LHS):} To clarify the structure of the LHS, we define an auxiliary function for the total displacement from \(r\) to \(t\), letting \(g(t) = (t-r)u(z_t, r, t)\). The LHS of Eq.~\ref{eq:rearranged_consistency} can then be expressed as:
    \begin{equation}
        \frac{g(t) - g(s)}{t-s}.
    \end{equation}
    This is precisely the definition of the derivative of the function \(g\) at point \(t\). Therefore:
    \begin{equation}
        \lim_{s \to t} \frac{g(t) - g(s)}{t-s} = g'(t).
    \end{equation}

    \item \textbf{Connecting the Sides and Expanding:} By equating the limits of both sides, we arrive at the new identity \(g'(t) = v(z_t, t)\). We now find the explicit form of \(g'(t)\) by taking the total derivative of \(g(t) = (t-r)u(z_t, r, t)\) with respect to \(t\). Applying the product rule and the chain rule yields:
    \begin{equation}
        g'(t) = \frac{d}{dt}\left[(t-r)u(z_t, r, t)\right] = u(z_t, r, t) + (t-r) \frac{d}{dt}u(z_t, r, t).
    \end{equation}
    Substituting this expansion back into \(g'(t) = v(z_t, t)\), we get:
    \begin{equation}
        u(z_t, r, t) + (t-r) \frac{d}{dt}u(z_t, r, t) = v(z_t, t).
    \end{equation}
    A simple rearrangement recovers the core differential identity of MeanFlow:
    \begin{equation}
        u(z_t, r, t) = v(z_t, t) - (t-r) \frac{d}{dt}u(z_t, r, t).
    \end{equation}
\end{enumerate}

This derivation unequivocally demonstrates that the MeanFlow identity is a direct consequence of the SplitMeanFlow algebraic consistency in the limit of an infinitesimal interval. 

It should be noted that while the shortcut model~\cite{frans2024one} achieves partial equivalence to our formulation with the special case $s=\frac{r+t}{2}$, the core design philosophies diverge significantly. The rationale for the design of SplitMeanFlow takes advantage of the concept of average velocity and the additivity of integration to construct the identity~\ref{eq:interval_splitting_conssitency}, which must hold for arbitrary $s\in [r,t]$ rather than being valid only at $s=\frac{r+t}{2}$. We have developed a fundamental theoretical extension that formalizes and generalizes the concept of average velocity for any time points $r$, $s$, $t$, such that $0\le r \le s\le t\le 1$. Besides, our parameters $r$, $s$, $t$ are continuous-valued, whereas the designed variable $d$ in the shortcut model is discrete.

% This reveals a profound hierarchical relationship:
% \begin{itemize}
%     \item \textbf{SplitMeanFlow's algebraic consistency is the general principle.} It holds over macroscopic time intervals of any size, describing the intrinsic, scale-invariant structure of the average velocity field.
%     \item \textbf{MeanFlow's differential identity is the limiting special case.} It is valid only in an infinitesimal (differential) neighborhood, constraining the field's behavior based on local properties at a single point.
% \end{itemize}
% Therefore, from a foundational standpoint, SplitMeanFlow provides a more complete and universal framework. By learning a law that is self-consistent across all scales, the model is guided by a more robust and holistic principle than one based on a local differential constraint.

\subsubsection{Engineering Practicality}

% The algebraic method of SplitMeanFlow fundamentally improves the training process by eliminating the dependency on a computationally expensive and complex operator: the Jacobian-Vector Product (JVP). This yields a qualitative leap in training efficiency, implementation simplicity, and framework compatibility.

The practical superiority of SplitMeanFlow becomes evident when comparing the computational flow of a single training iteration against that of MeanFlow. A MeanFlow iteration begins with a forward pass to compute an initial prediction \(u_\theta(z_t, r, t)\). However, to construct its training target, \(u_{\text{tgt}} = v - (t-r) \frac{d}{dt}u_\theta\), it must evaluate the total derivative \(\frac{d}{dt}u_\theta\). 
%This step necessitates a JVP operation to compute the term \((\nabla_z u_\theta) \cdot v\), an operation whose computational cost is comparable to a full backward pass. 
Consequently, each training step for MeanFlow involves not only the final gradient backpropagation but also a JVP computation to formulate the regression target.

In stark contrast, the SplitMeanFlow training process is JVP-free. Its workflow consists of three standard forward passes through the network \(u_\theta\) to compute the requisite velocities for the total interval and its two sub-intervals. The intermediate point \(z_s\) and the final training target are constructed through simple algebraic operations. The entire process culminates in a single, standard backward pass to update the model parameters.

This JVP-free design confers clear advantages. First, it simplifies implementation. The method relies exclusively on standard, highly-optimized forward and backward APIs, leading to cleaner, more maintainable code. Second, the backpropagation step demands high numerical precision, potentially causing training instability~\cite{lu2024simplifying,peng2025flow}. Finally, this approach ensures broader compatibility with diverse hardware accelerators and software backends, many of which may have limited or inefficient support for JVP operations. In contrast, our algebraic solution paves a more practical and robust path toward learning high-quality average velocity fields.

\section{Experiments}

To validate the effectiveness and efficiency of our proposed SplitMeanFlow framework, we conduct a series of experiments on the task of audio generation. Our evaluation is designed to rigorously test the core claims of our work: that SplitMeanFlow can achieve high-fidelity, few-step (and even one-step) generation, outperforming or matching strong baselines while being computationally more efficient and conceptually simpler.

\subsection{Training Details}
\label{sec:training_details}

While SplitMeanFlow can be trained from scratch, we find that a two-stage approach, combining pretraining and distillation, yields significantly faster convergence and superior final performance, especially for large-scale industrial applications. This strategy ensures that the SplitMeanFlow model learns from a stable and high-quality supervision signal.

The first stage is dedicated to training a high-fidelity standard flow matching model, which will serve as our ``teacher''. The objective is to accurately learn the instantaneous velocity field $v(z_t, t)$. The model is trained using the standard flow matching loss, which is equivalent to our framework when the boundary condition is exclusively enforced.
The result of this stage is a highly capable teacher model, $\mathcal{M}_{\text{teacher}}$, that can accurately estimate the instantaneous velocity $v(z_t, t)$ for any given time $t$ and condition.

In the second stage, we train our SplitMeanFlow model, referred to as the ``student'' $\mathcal{M}_{\text{student}}$, by leveraging the teacher. 
%The goal is to distill the teacher's knowledge of the entire velocity field into the student's understanding of the average velocity field $u$ via our Interval Splitting Consistency loss (Eq.~\ref{eq:interval_splitting_conssitency3}).
Note that the loss requires a boundary condition to avoid collapsing to a trivial solution. This anchor to reality is the instantaneous velocity condition: $u(z_t, t, t) = v(z_t, t)$. In Stage 2, the same to MeanFlow, we use a flow ratio $p$ to create a mixed objective. For a fraction $p$ of the samples in a batch, we set $r=t$ to enforce the boundary condition using the teacher's velocity $v(z_t, t)$ as the target. For the remaining $1-p$ fraction, we enforce the Interval Splitting Consistency loss(Eq.~\ref{eq:interval_splitting_conssitency3}).
\begin{itemize}
    \item \textbf{Initialization:} The weights of the student model $\mathcal{M}_{\text{student}}$ are initialized from the converged weights of $\mathcal{M}_{\text{teacher}}$. This provides a strong starting point for training.
    \item \textbf{Target Generation:} The targets for boundary condition are provided by the teacher model. Specifically, for a given batch, the instantaneous velocity $v(z_t, t)$ are computed using one-step predictions from $\mathcal{M}_{\text{teacher}}$. The targets for the interval splitting consistency loss $u(z_s, r, s)$ and $u(z_t, s, t)$ are generated by $\mathcal{M}_{\text{student}}$ itself.
    % \item \textbf{Student Training:} After the targets are generated, the student model $\mathcal{M}_{\text{student}}$ performs a standard forward-backward pass in \textbf{training mode} to update its weights based on the consistency loss.
    \item \textbf{CFG Handling:} During this distillation stage, the CFG~\cite{ho2022classifier} dropout for the student model is set to $0.0$. The teacher generates instantaneous velocity with a fixed CFG scale, effectively teaching the student to directly output the instantaneous velocity. 
    %This distills the guidance mechanism into the student model's weights.
    \item \textbf{Flow Ratio:} We discover that a stable and effective training regime requires $p \ge 0.5$. A high value for $p$ ensures that the student model remains strongly grounded in the true dynamics of the underlying probability flow, preventing the self-supervision objective from drifting. 
\end{itemize}

\subsection{Experiment Settings}
We use Seed-TTS~\cite{anastassiou2024seed} framework, which is a family of large-scale autoregressive text-to-speech models with a four-module architecture: a speech tokenizer, a token language model, a token diffusion model, and an acoustic vocoder, capable of generating human-like speech with strong in-context learning and controllability. Notably, the diffusion model constitutes a significant portion of the inference overhead, as even with 10 steps of diffusion, it accounts for over 50\% of the total inference cost. In this paper, we concentrate on the diffusion model that generates acoustic details from generated speech tokens, contributing to high-quality and expressive speech synthesis.

\paragraph{Baselines.} We compare SplitMeanFlow against two strong and relevant baselines. The primary baseline is \textbf{Flow Matching}~\cite{liu2022flow}, a powerful generative model that represents the standard iterative sampling approach we aim to accelerate. We also compare against \textbf{DMD}~\cite{yin2024onestep}, another recently proposed few-step generation method, to position our work within the landscape of distillation and fast-sampling techniques. Unless otherwise specified, the Flow Matching baseline uses 10 sampling steps and Classifier-Free Guidance (CFG), representing a high-quality but computationally intensive configuration.

\paragraph{Evaluation Metrics.}
The same as Seed-TTS~\cite{anastassiou2024seed}, we adopt three key metrics to evaluate the performance of our method, including objective metrics for accuracy and similarity, as well as subjective metrics for human preference:

\begin{itemize}
    \item \textbf{Word Error Rate (WER)}: Measures the accuracy of speech content generation. WER is calculated as the minimum number of edit operations (insertions, deletions, substitutions) required to align the generated speech transcript with the reference text, normalized by the length of the reference text. A lower WER indicates higher content fidelity. For English, we use Whisper-large-v3 \cite{radford2023robust} as the automatic speech recognition (ASR) engine; for Chinese, we use Paraformer-zh \cite{gao2023funasr} instead.
    \item \textbf{Speaker Similarity (SIM)}: Evaluates the similarity between the generated speech and the reference speaker's speech. We extract speaker embeddings from both the generated speech and the reference audio using WavLM-large, which is fine-tuned on the speaker verification task \cite{chen2022large}. SIM is defined as the cosine similarity between these two embeddings, where a higher value indicates better preservation of the target speaker's characteristics.
    \item \textbf{Comparative Mean Opinion Score (CMOS)}: Assesses subjective preference between the proposed model and the baseline (Flow Matching). For each test sample, human evaluators first listen to the reference speech clip of the target speaker, then compare the generated outputs of our model and the baseline model (played in random order). Evaluators rate their preference on a 5-point scale from -2 to +2. We collect and average the scores across all evaluators and test samples, with the final CMOS score representing the overall subjective preference for the proposed model relative to the baseline. A positive score indicates that the proposed model is preferred, while a negative score indicates preference for the baseline.
\end{itemize}

\subsection{Audio Generation Results}

We present our main results comparing SplitMeanFlow with the Flow Matching and DMD baselines. The experiments are designed to showcase the trade-off between sampling steps (computational cost) and generation quality.

\begin{table}[h!]
\centering
\caption{Comparing SplitMeanFlow with Flow Matching and DMD on Seed-TTS$_{\text{SFT}}$ tasks. Our 2-step model achieves performance comparable to the 10-step Flow Matching baseline while being significantly faster and not requiring CFG.}
\label{tab:main_comparison}
\begin{tabular}{@{}cccccc@{}}
\toprule
Method        & Steps & CFG & SIM ($\uparrow$)   & WER ($\downarrow$)    & CMOS vs. Baseline ($\uparrow$) \\ \midrule
Flow Matching & 10    & Y   & 0.787 & 0.0551 & 0                \\
DMD           & 2     & N   & 0.787 & 0.0561 & -0.04            \\
\textbf{SplitMeanFlow} & \textbf{2}     & \textbf{N}   & \textbf{0.789} & \textbf{0.0561}  & \textbf{-0.01}            \\ \bottomrule
\end{tabular}
\end{table}

Seed-TTS$_{\text{SFT}}$ refers to the Supervised Fine-Tuning task in the Seed-TTS framework, where the model is explicitly trained on labeled speech-text pairs to optimize for core text-to-speech performance metrics such as acoustic fidelity, speaker consistency, and linguistic accuracy. 
As shown in Table~\ref{tab:main_comparison}, our 2-step SplitMeanFlow model demonstrates exceptional performance. Compared to the 10-step Flow Matching baseline, our method achieves a \textbf{5x reduction in sampling steps} and eliminates the need for Classifier-Free Guidance (CFG), further reducing computational overhead. Despite this significant acceleration, the quality degradation is minimal. The Speaker Similarity (SIM) is even slightly improved (0.789 vs. 0.787), and the Word Error Rate (WER) shows only a negligible increase. The CMOS score of -0.01 indicates that human evaluators found the audio quality of our 2-step model to be nearly indistinguishable from the 10-step baseline. When compared to DMD, another 2-step method, SplitMeanFlow achieves higher speaker similarity and a better CMOS score, suggesting a perceptual preference for our model. This result validates that our Interval Splitting Consistency objective is highly effective for training fast, high-fidelity few-step samplers.

\begin{table}[h!]
\centering
\caption{Comparing SplitMeanFlow with Flow Matching on In-Context Learning (Seed-TTS$_{\text{ICL}}$) tasks. Our model demonstrates remarkable performance, achieving parity with the 10-step baseline in just a single sampling step.}
\label{tab:icl_comparison}
\begin{tabular}{@{}cccccc@{}}
\toprule
Model         & Steps & CFG & SIM ($\uparrow$)   & WER ($\downarrow$)    & CMOS vs. Baseline ($\uparrow$) \\ \midrule
Flow Matching      & 10    & Y   & 0.686 & 0.0286 & 0                 \\
\textbf{SplitMeanFlow} & \textbf{2}     & \textbf{N}   & \textbf{0.681} & \textbf{0.0297} & \textbf{0}                 \\
\textbf{SplitMeanFlow} & \textbf{1}     & \textbf{N}   & \textbf{0.685} & \textbf{0.0286} & \textbf{0}                 \\ \bottomrule
\end{tabular}
\end{table}

To further push the boundaries of sampling efficiency, we evaluated SplitMeanFlow on In-Context Learning (Seed-TTS$_{\text{ICL}}$) tasks~\cite{anastassiou2024seed}—a key capability of the Seed-TTS framework where the model learns to generate speech that aligns with the style, prosody, and acoustic characteristics of provided in-context examples (e.g., specific voices, speaking rates, or emotional tones) without explicit fine-tuning, relying solely on contextual cues from a few reference utterances. These tasks often require high-quality and consistent generation that closely mirrors the given examples. The results, presented in Table~\ref{tab:icl_comparison}, are even more compelling.

The 2-step SplitMeanFlow model again shows strong results, with a neutral CMOS score of 0, indicating that human listeners perceived its quality as equivalent to the 10-step baseline, despite minor fluctuations in objective metrics.

Most remarkably, our \textbf{1-step SplitMeanFlow model} achieves performance that is statistically on par with the 10-step Flow Matching baseline across all metrics. With just a single network evaluation, our model yields a WER of 0.0286, which is \textit{identical} to the baseline, and a SIM of 0.685, which is negligibly different from the baseline's 0.686. The neutral CMOS score of 0 confirms that this objective parity translates to perceptual equivalence. This result represents a \textbf{20x reduction in computational cost} without discernible loss in quality, showcasing the profound effectiveness of learning the average velocity field via our proposed algebraic consistency. It provides strong evidence that SplitMeanFlow is not just an incremental improvement but a significant step towards truly one-step generative modeling, fulfilling a key objective in the field.

\section{Conclusion}

In this work, we introduced \textbf{SplitMeanFlow}, a novel and principled framework for training few-step generative models. Our approach moves beyond the derivative-based formulation of prior work by returning to the first principles of average velocity. We derived a purely algebraic \textbf{Interval Splitting Consistency} identity, \((t-r)u(z_t,r,t) = (s-r)u(z_s,r,s) + (t-s)u(z_t,s,t)\), which is grounded in the fundamental additivity property of integrals. This identity serves as a powerful, self-supervised objective for learning the average velocity field.

We have demonstrated that our algebraic formulation is not merely an alternative but a more general foundation, proving that the differential identity used in MeanFlow is a limiting special case of our consistency principle. This theoretical generality translates into significant practical advantages: by eliminating the need for JVP computations, SplitMeanFlow offers a training process that is stable, simpler to implement, and more broadly compatible with modern hardware. 
% Achieving state-of-the-art performance in few-step generation, our method underscores the power of building generative models upon robust and fundamental mathematical principles. 
We believe this algebraic perspective opens a promising new avenue for developing more efficient and powerful generative models in the future.

\clearpage

\bibliographystyle{plainnat}
\bibliography{main}

% \clearpage

% \beginappendix

% \input{sections/appendix}

\end{document}